\documentclass[10pt,journal,compsoc]{IEEEtran}
\usepackage{mathrsfs}
\usepackage{adjustbox}
\usepackage{multirow}
\usepackage{amssymb} 
\usepackage{subfigure}
\usepackage{algorithm}
\usepackage{color}
\usepackage{algorithmic}

\usepackage[switch]{lineno}

\ifCLASSOPTIONcompsoc
  \usepackage[nocompress]{cite}
\else
  \usepackage{cite}
\fi
\ifCLASSINFOpdf
\else
\fi
\begin{document}
%
\title{Neuron Coverage-Guided Domain Generalization}
%
%
%
%

\author{Chris Xing Tian,~
        Haoliang Li,~
        Xiaofei Xie,~
        Yang Liu,~
        and~ Shiqi Wang
\IEEEcompsocitemizethanks{

\IEEEcompsocthanksitem C.X.Tian and S.Wang are with the Department of Computer Science, City University of Hong Kong, Hong Kong SAR, China \protect\\
\IEEEcompsocthanksitem H. Li is with the Department of Electrical Engineering, City University of Hong Kong, Hong Kong SAR, China.\protect\\

\IEEEcompsocthanksitem X. Xie is with the School of Computing and Information Systems, Singapore Management University, Singapore \protect\\

\IEEEcompsocthanksitem Y. Liu are with the School of Computer Science and Engineering, Nanyang Technological University, Singapore.\protect\\

}
}

\markboth{Journal of \LaTeX\ Class Files}%
{Shell \MakeLowercase{\textit{et al.}}: Bare Demo of IEEEtran.cls for Computer Society Journals}
%



\IEEEtitleabstractindextext{%
\begin{abstract}

This paper focuses on the domain generalization task where domain knowledge is unavailable, and even worse, only samples from a single domain can be utilized during training. Our motivation originates from the
recent progresses in deep neural network (DNN) testing, which has shown that maximizing neuron coverage of DNN can help to explore possible defects of DNN (i.e., misclassification). 
More specifically, by treating the DNN as a program and each neuron as a functional point of the code, during the network training we aim to improve the generalization capability by maximizing the neuron coverage  of DNN with the gradient similarity regularization between the original and augmented samples. 
As such, the decision behavior of the DNN is optimized, avoiding the arbitrary neurons that are deleterious for the unseen samples, and leading to the trained DNN that can be better generalized to out-of-distribution samples. Extensive studies on various domain generalization tasks based on both single and multiple domain(s) setting demonstrate the effectiveness of our proposed approach compared with state-of-the-art baseline methods. We also analyze our method by conducting visualization based on network dissection. The results further 
provide useful evidence on the rationality and effectiveness of our approach. 
\end{abstract}

}

\maketitle

\IEEEdisplaynontitleabstractindextext

%
\IEEEpeerreviewmaketitle

\IEEEraisesectionheading{\section{Introduction}\label{sec:introduction}}

It has been well-known that training a deep neural network (DNN) model with desirable generalization ability usually requires abundant labeled data~\cite{zhang2016understanding,kawaguchi2017generalization}, and similar to other supervised learning models, the trained deep model may fail to generalize well on the out-of-distribution samples which are different from the domains where it is trained~\cite{DBLP:conf/nips/YosinskiCBL14}. To overcome these limitations, transfer learning, which aims to transfer knowledge from some source domain(s) to boost the generalization performance of a learning model on a domain of interest (i.e., a target domain), has been proposed~\cite{pan2010survey}.

However, in many applications, it may not be feasible to acquire the target domain data in advance. Domain generalization \cite{muandet2013domain} is a promising technique which gains knowledge acquired from different domains, and applies it to previously unseen but related domains.  Generally speaking, current research regarding domain generalization can be categorized into two
streams. The first stream (e.g., \cite{muandet2013domain}) aims at learning to extract a universal feature representation among domains
 through either distribution alignment or multi-task learning. The other stream (e.g., \cite{li2017learning}) leverages
the advantage of meta-learning methods for feature representation learning which was originally proposed for few-shot learning problem. 
During training, meta-train set and meta-test set are
selected from source domains during each iteration to produce a more robust feature representation.

One major limitation of aforementioned methods lies in that the prior knowledge of domain information must be acquired in advance, such that one can leverage covariate shift \cite{torralba2011unbiased} to learn shareable information among domains.  In practice, the training data can be quite complicated that there may not be known and clear distinction among the domains, thus, the domain of each sample is ambiguous to define. Even worse, due to the privacy issue, one may only have data collected from one single domain, 
such that it is impractical to simulate covariate shift with large domain gap in this case. 

To tackle the aforementioned problem, there exists works focusing on the worst-case formulation of domain generalization that the samples from only one single domain can be utilized during training stage. The main idea is conducting data augmentation to improve the generalization capability of DNN. In \cite{volpi2018generalizing}, an adversarial training mechanism was proposed by forcing the latent features of the original and augmented data to be lying on a similar manifold. Such mechanism has been further extended in \cite{qiao2020learning}, where a Wasserstein distance constraint \cite{tolstikhin2017wasserstein} was introduced to encourage out-of-distribution augmentation and a meta-learning method was adopted to learn shareable information between the original and augmented data. 
 
The goal of these works is to guarantee that the original data and its augmented data have the similar prediction/decision by the DNN. However, how to understand the misclassification behavior of an out-of-distribution sample and how to measure the similarity of the decision on two inputs are still open problems.

The coverage analysis has been recognized as the key driver in software testing, ensuring no bugs triggered in the high-coverage tests. Software bugs usually trigger the abnormal behaviors that cover the special control/data flow of the programs. In the traditional code testing, software bugs can be regarded as the out-of-distribution inputs which have different coverage compared with normal inputs. Inspired by this, the neuron coverage was further proposed to detect out-of-distribution samples that may have different coverage, which leads to the misclassification behavior of DNNs \cite{pei2017deepxplore,ma2018deepgauge}. 
The undetected bugs (i.e., misclassification) of DNN can be caused by the neurons which are inactivated during training stage but can be activated by the out-of-distribution samples during testing stage \cite{pei2017deepxplore}. Thus, in order to improve the generalization capability of the DNN, we aim to maximize the coverage such that more neurons can be activated during the training phase. Moreover, in analogous to the traditional software, the out-of-distribution samples and the in-distribution samples should trigger the similar control flow or data flow of the DNN if they have similar semantic information or logic (i.e., same label information for recognition task). 

To this end,  we propose to improve the generalization capability of DNNs by \textit{maximizing the neuron coverage} of DNN with \textit{gradient similarity regularization} between the original and the augmented samples, where neuron coverage and gradient can represent the control flow and data flow of DNN, respectively. 
We expect the trained DNN to be better generalized to out-of-distribution samples from unseen but related domains. Experimental results on domain generalization setting where domain knowledge is not available during the training stage (including the worst-case formulation where only one single domain is available) demonstrate the effectiveness of our proposed method. Last but not least, we also attempt to bridge the gap between the software testing and computer vision by providing analysis based on network visualization through network dissection \cite{bau2020understanding}. The visualization results further justify the rationality and the effectiveness of our proposed method.

\section{Related Works}
\subsection{Domain Generalization}
The goal of domain generalization is to improve the generalization capability of the trained model towards the out-of-distribution samples from unseen but related domains. One main research direction of domain generalization is to learn shareable representations from samples of different domains. For example,  Muandet \textit{et al.} \cite{muandet2013domain} proposed the Domain Invariant Component Analysis (DICA) algorithm based on multiple source-domain data. As such, the distribution mismatch across domains is minimized while the conditional function relationship is preserved. Ghifary~\textit{et al.}~\cite{ghifary2015domain} proposed a multi-task autoencoder framework to learn domain invariant features by reconstructing the data from one domain to another. Motiian~\textit{et al.}~\cite{Motiian_2017_ICCV} proposed to minimize the semantic alignment loss as well as the separation loss for domain generalization. Li~\textit{et al.}~\cite{li2017deeper} proposed to learn CNN model through low-rank regularization. Carlucci~\textit{et al.}~\cite{carlucci2019domain} proposed to leverage the advantage of self-supervised learning to learn generalized feature representation. Wang~\textit{et al.}~\cite{wang2019learning} proposed to learn robust feature representation through statistics out by removing the grey-level co-occurrence information. More recently, Zhou~\textit{et al.}~\cite{zhou2020deep} proposed to conduct image generation across domains. Huang~\textit{et al.} \cite{huang2020self} proposed a ``dropout on gradient" mechanism for domain generalization.  

Another direction is to conduct meta-learning (a.k.a. learning to learn) by simulating domain shift for meta-train and meta-test set to tackle the problem of domain generalization. In \cite{li2017learning}, the idea of meta learning, which was originally proposed for few-shot learning problem, was extended to the ``unseen" domain setting. Balaji \textit{et al.} \cite{balaji2018metareg} proposed an episodic training procedure by considering regularization based on domain specific network. Such idea was further extended with either advanced training mechanism \cite{li2019episodic}, novel network regularization \cite{li2019feature} or additional feature embedding loss (e.g.,  triplet loss) \cite{dou2019domain}.

While the aforementioned techniques focusing on multiple domain setting, recently, the worst-case formulation of domain generalization that only single domain training data can be utilized has also drawn more and more attentions.  In \cite{volpi2018generalizing}, an adversarial training mechanism was proposed with the regularization that augmented data and the original data lie on a similar manifold in terms of the semantic feature space. In \cite{qiao2020learning},  a joint data augmentation method was proposed based on adversarial training mechanism 
for within-domain data augmentation. Moreover, the Wasserstein autoencoder for out-of-manifold data augmentation has been presented in  \cite{tolstikhin2017wasserstein}. Self-supervised learning has also been proved to be effective in this case.  For example, in \cite{carlucci2019domain}, the authors have shown that better generalization capability can be achieved based on single domain scenario by performing the evaluations in 
the digit recognition task. 

\subsection{Code Coverage and Neuron Coverage}
In traditional program, control-flow graph (CFG) is a representation of all paths that might be executed. Each node of the CFG is a basic block including a sequence of statements. The edge represents the jump (e.g, the if-else branch) from one basic block to another one. Data flow analysis aims to gather information regarding the possible set of values calculated at various points of the CFG. To test the programs, 
several coverage criteria (e.g., statement coverage, branch coverage) have been proposed based on the CFG. Motivated by the structure of traditional programs, Pei \textit{et al.} \cite{pei2017deepxplore} proposed the neuron coverage that measures the percentage of activated neurons with given input set. Furthermore, Ma \textit{et al.} \cite{ma2018deepgauge} then extended neuron coverage and proposed a set of more fine-grained neuron-based coverage criteria considering the distribution of the neuron outputs from training data. Unlike the aforementioned techniques which applied neuron coverage during the DNN testing stage, we propose to maximize neuron coverage during the training stage for improving the generalization capability to unseen domains. 
\section{Methodology}

\subsection{Preliminary and Overview}

We denote the source domain(s) on a joint space $\mathcal{X} \times \mathcal{Y}$ as $\mathcal{D}_S = \{\mathbf{x}_l^{S},y_{l}^S\}_{l=1}^{N_S}$ with $N_S$ labeled samples in total. We aim to learn a DNN model $f$ which is parameterized by $\Theta$ with $\mathcal{D}_S$ only, and perform the classification task on the samples $\mathbf{x}^T$ from a related domain $\mathcal{D}_T$ without having any prior observations or knowledge about $\mathcal{D}_T$.  

We provide a framework called Neuron Coverage-guided Domain Generalization (NCDG) to improve the generalization of DNN model $f$. 

\begin{figure}[t]
\centering
\includegraphics[width=0.8\columnwidth]{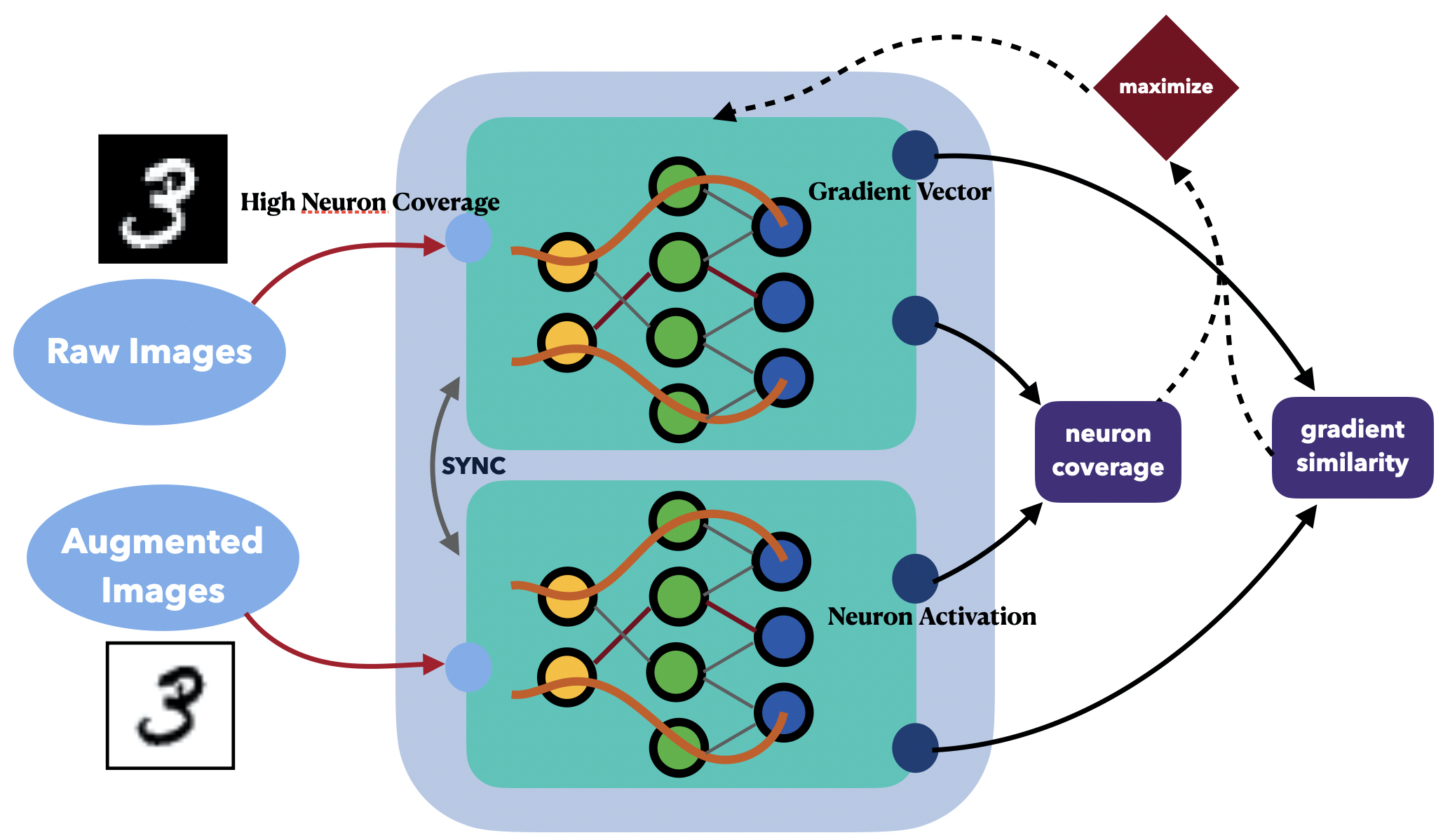}
\caption{Our proposed framework for domain generalization. Given the two different samples with similar semantics, we propose to maximize the neuron coverage of DNN with gradient similarity regularization between two samples with similar semantic information.}
\label{fig:framework}
\vspace{-5pt}
\end{figure}

Inspired by the CFG and data flow graph (DFG) in traditional programs, we propose novel loss functions, i.e, the \textit{neuron coverage maximization loss} with the \textit{gradient similarity loss} regularization.
By optimizing the proposed loss function on inputs from two  similar domains (i.e., the original source domain $\mathcal{D}_S$ and its augmented domain $\hat \mathcal{D}_S$), we expect the DNN to be better generalized to out-of-distribution samples. Herein, we  introduce the two loss functions as well as the training procedure. The whole framework is summarized in Fig.~\ref{fig:framework}.


\subsection{Neuron Coverage Maximization}
Existing works~\cite{pei2017deepxplore,ma2018deepgauge} have demonstrated that the out-of-distribution samples 
exhibit different neuron coverage with the original data (i.e., different neurons are activated). To mitigate this issue, we propose to maximize the neuron coverage of the original data and the augmented data based on the following inspirations. First, maximizing the neuron coverage can improve the prediction stability by reducing the possibility that the out-of-distribution samples activate different neurons with the training data. We expect that the undetected bugs (i.e., misclassification)  caused by inactive neurons, which are triggered by out-of-distribution samples during testing stage, can be reduced.  Second, by maximizing the neuron coverage, we can further expect more overlapped activated neurons between the original data and the augmented data. As such, the neuron activation similarity between 
the original data and the augmented data is increased (i.e., the ``control flow'' is similar).

We denote $n_i^j$ as the $j$-th neuron in the $i$-th layer of DNN $f$, where the output of $n_i^j$ is denoted as $out(\mathbf{x},n_i^j)$ given input $\mathbf{x}$. We further conduct max-min normalization of $out(\mathbf{x},n_i^j)$ given the neuron outputs from the same layer 
$\mathcal{O}(\mathbf{x},i) = \{out(\mathbf{x},n_i^1),out(\mathbf{x},n_i^2),...,out(\mathbf{x},n_i^{N_i})\}$, where $N_i$ is the total number of neurons in the $i$-th layer. The normalized output of $out(\mathbf{x},n_i^j)$ is represented as
\begin{equation}
\small
\tilde{out}(\mathbf{x},n_i^j) = \frac{out(\mathbf{x},n_i^j) - min(\mathcal{O}(\mathbf{x},i))}{max(\mathcal{O}(\mathbf{x},i)) -  min(\mathcal{O}(\mathbf{x},i))}.
\end{equation}
As such, the neuron $n_i^j$ is considered to be activated only if $\tilde{out}(\mathbf{x},n_i^j)$ is larger than a pre-defined threshold $t$.
 \begin{algorithm}
\caption{The Computation of Neuron Coverage Loss}
\label{alg:A}
\begin{algorithmic}[1] 
\REQUIRE ~~\\ 
\textbf{$\lambda$} $\to$ parameter for neuron coverage;\\
{$t$} $\to$ threshold for neuron activation;\\
\textbf{model} $\to$ model for training;\\
\textbf{neuron\_act\_map} $\to$ records the neurons which have been ever activated, and it is initialized as empty map before the epoch.\\
\ENSURE ~~\\
\textbf{neu\_cov\_loss} $\to$ neuron coverage loss for current ended iteration, to be maximized in following iteration;\\

\STATE /*\ $one \ training\ iteration\ starts$\ */
\STATE neu\_cov\_loss := 0 
\FOR{layer \textbf{in} model.layers}
\STATE inactive\_neurons = \{\}
\IF{layer \textbf{not}  activation layer}
\STATE continue 
\ENDIF
\STATE max = maximal output value \textbf{in} layer.neurons \\
\STATE min = minimal output value \textbf{in} layer.neurons \\
\FOR{ $n$ \textbf{in} layer.neurons}
\STATE n\_val\_scale = ($n$.output() - min) /  (max - min)
\IF{ neuron\_act\_map($n$) \textbf{or} n\_val\_scale \textgreater t}
\STATE neuron\_act\_map($n$) = $\TRUE$
\ELSE
\STATE inactive\_neurons.put($n$, n\_val\_scale)
\ENDIF
\ENDFOR
\STATE neu\_cov\_loss += sum(inactive\_neurons.outputs())
\ENDFOR
\\
\STATE \textbf{return} neu\_cov\_loss /*$to\ be\ incorporated\ with\ \mathcal{L}_{c}$\ */
\end{algorithmic}
\end{algorithm}

The neuron coverage is defined based on the proportion of neurons which have been activated given a batch of data~\cite{pei2017deepxplore,ma2018deepgauge}. Moreover, the neuron is considered to be activated if there exists a sample from the batch where the corresponding output is larger than the threshold $t$. 

In analogy to the implication of low code coverage in software testing, low neuron coverage of a DNN may suggest that the incorrect DNN behaviors (i.e., misclassification) remains unexplored. Thus, in this scenario the DNN tends to be lack of generalization capability as errors can be induced once the inactivated neurons get activated during the test phase on target domains.

Directly maximizing the neuron coverage of DNN based on a batch of data may not be computationally feasible due to the \textit{logical reasoning} involved \cite{serafini2016logic}. To this end, we first propose to relax the optimization by maximizing the average output of neuron given a batch  together with the minimizing the standard classification loss (e.g., cross-entropy loss) $\mathcal{L}_{c}(f,\mathcal{D})$ to improve the generalization capability of DNN, which can be represented as 
\begin{equation}\label{cov_loss}
\small
\mathcal L_{cov}(f,\mathcal{D}) = \mathcal{L}_{c}(f,\mathcal{D}) - \lambda \mathbb{E}_{\mathbf{x} \in \mathcal{D}} \sum_{i}\sum_{j} {\tilde{out}(\mathbf{x},n_i^j;t)},
\end{equation}
where $\mathcal{D}$ denotes the domain for training, $\lambda$ is the parameter for neuron activation loss, and $t$ is the threshold for neuron activation. 

We further propose a bootstrapping-like mechanism to optimize Eq.~\ref{cov_loss} by exploring the \textit{inactive neurons} for more efficient training.  More specifically, we first initialize the inactivated set of the $i$-th layer $\mathcal {IS}_{i}$ to contain all neurons from the output of this layer at the beginning of a training epoch. We then update the $\mathcal {IS}_{i}$ by removing the neurons which have been activated during optimization (i.e., the output of neuron is higher than the threshold $t$). Thus, in each iteration, we only consider the neurons in $\mathcal {IS}_{i}$  to compute average output of neurons instead of adopting all neurons from the $i$-th layer. We repeat the process until the end of the epoch, and reset all neurons to be inactivated again at the beginning of the next epoch accordingly. The details of computing the neuron coverage loss are summarized in Algorithm 1.

While the neuron coverage can be used to measure the decision similarity, it still cannot model the data flow through skip connections over certain layers (e.g., residue connection module~\cite{he2016deep}). We propose to adapt the \textit{gradient information} to represent the data flow of the DNN, as it can track the neuron output and the information flow across the shortcut \cite{he2016identity}.  More specifically, we use gradient of neuron coverage loss (i.e., standard classification loss with neuron activation maximization) for regularization purpose, which is denoted as $ \frac{\partial L_{cov}(f,\mathcal{D}_S)}{\partial \Theta} $ and can be obtained by optimizing $\mathcal{L}_{cov}$ through backpropagation.
Based on gradient information, we can measure the similarity between the original data and the augmented data in a finer-grained way.

Given the original source samples as $\mathcal{D}_S = \{\mathbf{x}_i^{S},y_{i}^S\}_{i=1}^{N_S}$, we first augment the data through data augmentation techniques to obtain $\hat \mathcal{D}_S$ (noted that $\hat \mathcal{D}_S$ and $\mathcal{D}_S$ are one-one correspondence). Subsequently, at each iteration, we train the model on the minibatch of $\mathcal{D}_S$ and $\hat \mathcal{D}_S$ separately with $\mathcal L_{cov}$ to obtain the gradients $ \frac{\partial L_{cov}(f,\mathcal{D}_S)}{\partial \Theta}$ and $ \frac{\partial L_{cov}(f,\hat \mathcal{D}_S)}{\partial \Theta}$, respectively. We model the gradient similarity between $\mathcal{D}_S$ and its augmented version $\hat \mathcal{D}_S$ based on L2 norm as
\begin{equation}\label{eq:sim}
\small
\mathcal L_{sim}(f) = \left\|\frac{\partial L_{cov}(f,\mathcal{D}_S)}{\partial \Theta} - \frac{\partial L_{cov}(f,\hat \mathcal{D}_S)}{\partial \Theta}\right\|_2,
\end{equation}
where the update of Eq.~\ref{eq:sim}  involves a gradient through a gradient and requires an additional backward pass through the model \cite{finn2017model}.

\textbf{Discussion: } It is worth noting that our proposed neuron coverage is computed based on the whole training dataset, which aims to increase the number of activated neurons (i.e., which have been activated by at least one sample from the training set). Our proposed method also does not contradict with dropout, as dropout is randomly applied on a batch (or a single sample). Our proposed gradient similarity regularization term is close to the contrastive learning (e.g., \cite{chopra2005learning,oord2018representation}) which aims to learn a representation by maximizing similarity and dissimilarity of data samples organized into similar and dissimilar pairs. Unlike these methods which map the data into a single (or multiple) embedding space(s) where contrastive learning is performed, our proposed method is based on the neuron coverage with similarity regularization by mapping the data into neural tangent kernel space \cite{jacot2018neural} (i.e., mapping the data from the original space to the kernel space represented by the gradient corresponding to the loss function). Our gradient similarity regularization is also different from \cite{li2017learning}. In \cite{li2017learning}, the gradient regularization is conducted from the perspective of meta-learning through first-order Taylor expansion based on average gradient through gradient descent, which requires the ``direction of improvement in each set of domains is similar". However, our proposed method is motivated by the data flow similarity, which requires gradient regularization in a one-to-one correspondence manner.  Last, while one may argue that removing inactivated neurons through network pruning can avoid unexpected activation, it has been shown in \cite{chen2019cooperative} that network pruning can even worsen the performance under cross-domain setting.  

\begin{algorithm}
\caption{Training Procedure of NCDG}
\label{alg:B}
\begin{algorithmic}[1] 
\REQUIRE ~~\\ 
\textbf{$\mathcal{D}_S$} $\to$ original training samples batch;\\
\textbf{$\hat \mathcal{D}_S$} $\to$ augmented training samples batch, one-one correspondence with $\mathcal{D}_S$;\\
\textbf{$\beta$} $\to$ parameter balancing coupled neuron coverage loss and gradient similarity loss;\\
\textbf{$\lambda, t$} $\to$ parameters for neuron coverage loss \\ 


\STATE /*\ $main\ procedure\ starts$\ */
\FOR{ite \textbf{in} iterations}
\STATE \textbf{Raw-train:} Gradients $\nabla_\Theta = \frac{\partial  \mathcal L_{cov}(f, \mathcal{D}_S)}{\partial \Theta}$
\STATE \textbf{Augmented-train:} Gradients $\hat \nabla_\Theta = \frac{\partial  \mathcal L_{cov}(f, \hat\mathcal{D}_S)}{\partial \Theta}$
\STATE \textbf{optimization:}  Update $\Theta$:
\STATE $\Theta = \Theta - \eta \cdot \frac{\partial \left( \mathcal L_{cov}(f, \mathcal{D}_S) + \mathcal L_{cov}(f,\hat\mathcal{D}_S) + \beta \cdot \left\| \nabla_\Theta - \hat \nabla_\Theta\right\|_2\right)}{\partial \Theta}$
\ENDFOR
\\
\STATE \textbf{end procedure}
\end{algorithmic}
\end{algorithm}

\subsection{Model Training}
In this subsection, we summarize the proposed method NCDG. In particular, the neuron coverage losses for $\mathcal{D}_S$ and $\hat \mathcal{D}_S$ are optimized, containing a standard classification loss with a neuron activation maximization regularization term. Moreover, a gradient similarity loss based on gradient between $\mathcal{D}_S$ and  $\hat \mathcal{D}_S$ is incorporated. As such, the final objective is given by
\begin{equation}
\small
\mathcal L_{NCDG} =  \mathcal L_{cov}(f, \mathcal{D}_S) + \mathcal L_{cov}(f, \hat \mathcal{D}_S) + \beta \mathcal L_{sim}(f, \mathcal{D}_S, \hat \mathcal{D}_S),  \label{loss_sim_cov}
\end{equation}
where $\beta$ is a parameter for balancing between neuron coverage loss and neuron gradient similarity loss. We also show the training details in Algorithm 2. 

It is worth mentioning that our proposed algorithm involves ``gradient through a gradient" operation, where the computational cost increases linearly  upon  the number of model layers. Nevertheless, our computational cost is still tractable according to the following analyses. 

\begin{enumerate}
    \item The outputs of neurons and gradients share the same size. By treating the neuron output and gradients as the input of the loss function, our proposed coverage loss shares the similar computational cost compared with the gradient-regularization based technique \textit{when conducting backward computation}.   
    \item In analogous to other domain adaptation techniques that conduct domain alignment on latent feature spaces, our proposed coverage loss does not necessarily need to maximize the coverage for  \textit{all} neurons. For example, while we consider all neurons in the Digit Recognition task, only the neurons in the convolutional layers are considered for Object Recognition task. Therefore, one 
   can further reduce the computational cost by only involving partial neurons for coverage maximization. How to select the neurons for coverage maximization will be investigated in our future work.  
    \item While the computational cost of our proposed method is higher than that of the vanilla model (i.e., directly training the model with cross-entropy loss), \textit{the computational cost of our method is conceptually the same as vanilla model during the inference stage}. 
\end{enumerate}

\section{Experiments}
\subsection{Single-Source Domain Generalization}

To evaluate the performance of NCDG in domain generalization, we first focus on a worst-case scenario, namely single-source domain generalization (SSDG), and compare our NCDG with state-of-the-art SSDG methods, including JiGen \cite{carlucci2019domain}, GUD \cite{volpi2018generalizing} and the recently proposed M-ADA \cite{qiao2020learning}, as well as the  Empirical Risk Minimization
(ERM) baseline \cite{koltchinskii2011oracle}.

\subsubsection{SSGD Evaluation on Digit Recognition}

We first evaluate on a standard SSDG benchmark \emph{Digits} \cite{volpi2018generalizing}. In particular, the model is trained on one single source dataset: MNIST \cite{lecun1998gradient} and tested on other four digital datasets including SVHN \cite{netzer2011reading}, MNIST-M \cite{ganin2015unsupervised}, SYN \cite{ganin2015unsupervised} and USPS \cite{denker1989neural} all at once. Following the experimental protocol of prior SSDG works including GUD \cite{volpi2018generalizing} and M-ADA \cite{qiao2020learning}, we select the first 10,000 images from MNIST train split for training, and then evaluate the classification accuracy on the test split of 
the four testing datasets as four different domains. All images are resized to $32 \times 32$. We also convert MNIST and USPS from grey scale to RGB image by duplicating the grey channel twice.

\begin{table}[h]
\begin{center}
\caption{SSDG classification accuracy (\%) on \emph{Digits}. The superscript $^{+}$ denotes the augmentation with our proposed loss. The superscript $^{**}$ denotes the baselines with pixel intensity reversing. In particular, the Vanilla$^{**}$ denotes the Vanilla scheme (only cross-entropy loss) with the pixel intensity reversing. By comparing the NCDG with the four schemes augmented with pixel intensity reversing (Vanilla$^{**}$, JiGen$^{**}$, GUD$^{**}$, M-ADA$^{**}$), it is apparent that the proposed loss achieves better performance under the same augmentation method. By comparing the GUD$^{+}$ and GUD, as well as the M-ADA$^{+}$ with M-ADA, significant performance improvement originating from our proposed loss is observed.  
}
\begin{adjustbox}{max width=\columnwidth}
\begin{tabular}{lccccc}
\hline
Method   & SVHN  & MNIST-M & SYN   & USPS  & Avg.  \\ \hline\hline
ERM  &  27.8 & 52.7 & 39.6 & 76.9 & 49.3 \\
JiGen    & 33.8 & 57.8   & 43.8 & 77.2 & 53.1 \\
GUD      & 35.5 & 60.4   & 45.3 & 77.3 & 54.6 \\
M-ADA    & 42.6 & 67.9   & 49.0 & 78.5 & 59.5 \\ \hline
GUD$^{+}$ & 40.3 & 62.1   & 46.4 & 80.3 & 57.3 \\
M-ADA$^{+}$   &{47.7}  &{70.7} &{51.9} &81.9 &{62.5} \\ \hline
Vanilla$^{**}$ &54.2 &73.9 &58.6 &90.9 &69.4 \\
JiGen$^{**}$ &53.7 &75.2 &60.1 &91.9 &70.2 \\
GUD$^{**}$ & 56.3 & 77.1   & 62.3 & 90.3 & 71.5 \\
M-ADA$^{**}$ &58.0 & \textbf{78.1} &60.9 &91.1 &72.0 \\\hline

{NCDG} & \textbf{59.7} &77.4   & \textbf{63.8} & \textbf{92.6} & \textbf{73.4} \\
\hline

\end{tabular}
\end{adjustbox}

\label{tab:table-digits}
\end{center}
\end{table}

 
Following the same protocol, a ConvNet with architecture \emph{conv-pool-conv-pool-fc-fc-softmax} is used as the training model. Adam optimizer with learning rate of 0.0001 and batch size of 32 is adopted. The model is trained for 32 epochs, which is equivalent to 10,000 iterations. For NCDG, we track the activation of neurons in all layers before the softmax and set activation threshold $t$ to 0.005 and other parameters as $\lambda = 0.1,  \beta = 0.01 $. Since original MNIST samples are grey-scale images, we choose to apply intensity reversing \cite{gonzalez2004digital} for augmentation purpose\footnote{It is worth mentioning that other augmentation methods can also be applied.}. To further 
validate the proposed method, we compare it with another baseline model (Vanilla$^{**}$) trained directly on original and augmented MNIST data without domain generalization.

 We report the results in Table \ref{tab:table-digits}. As we can observe, all domain generalization techniques can outperform the baseline model ERM, and NCDG can achieve significant performance improvement in all scenarios. On  the other hand, we can also achieve desired performance by directly training on the original and augmented data with cross-entropy loss (i.e., Vanilla$^{**}$). We conjecture that the reason lies in the intensity reversing which can enhance the structure information for digit recognition.  By involving the proposed domain generalization regularization term, we can achieve $4\%$  performance improvement on average compared with directly training on the original and augmented MNIST. 
 
 { We also compare the proposed method with the baseline schemes augmented by pixel intensity reversing (JiGen$^{**}$, GUD$^{**}$, M-ADA$^{**}$). As we can observe, while better performance can be achieved for baseline techniques with pixel intensity reversing, our proposed method can still better performance in most of the cases.
 
 Last but not the least, we justify that our method can be applied with other augmentation based methods. To this end, we consider the adversarial augmentation method proposed in GUD \cite{volpi2018generalizing} and M-ADA \cite{qiao2020learning}  and apply our proposed loss for training. We observe that the improvement can be achieved by a large margin compared with GUD and M-ADA, which further illustrates the  effectiveness of our proposed method.  }


\subsubsection{SSDG Evaluation on PACS}

 We consider a more realistic dataset: {PACS} \cite{li2017deeper} for evaluation. {PACS} covers objects of 7 different classes: \emph{dog, elephant, giraffe, guitar, house, horse, person} and consists of 4 domains including \emph{Art-Painting}, \emph{Cartoon}, \emph{Photo} and \emph{Sketch}. For SSDG evaluation, the images for training are from a single domain, and the performance of the model is evaluated on another domain.

\begin{table}[]
\caption{SSDG classification accuracy (\%) on PACS dataset with Resnet-18 as backbone model. Each row indicates the result of training on a single source domain and testing on the other three domains.}
\begin{adjustbox}{max width=\columnwidth}
\begin{tabular}{c|lcccc|c}
\hline
Source Domain                  & Method   & Photo & Art\_painting & Cartoon & Sketch & Avg. \\  \hline\hline
\multirow{5}{*}{Photo}         
& DeepAll   &/ &66.0 &26.7 &35.0 &42.5 \\
                               & JiGen    &/ &64.1 &23.9 &32.9  &40.3 \\
                               & GUD      &/ &55.8 &\textbf{33.3} &45.6  &44.9 \\
                               & M-ADA    &/ &64.3 &29.8 &35.2  &43.1 \\ \cline{2-7} 
                               & NCDG &/ &\textbf{68.8} &29.8 &\textbf{48.6} &\textbf{49.0} \\ \hline\hline

\multirow{5}{*}{Art\_painting} 
& DeepAll  &\textbf{96.5} &\ &60.6 &52.5 &69.9 \\
                               & JiGen    &{95.5} &/ &60.1 &50.2 &68.6 \\
                               & GUD      &93.7 &/ &61.1 &56.2 &70.4 \\
                               & M-ADA    &95.0 &/ &61.5 &47.6 &68.0 \\ \cline{2-7} 
                               &  NCDG &95.0 &/ &\textbf{68.6} &\textbf{66.4} &\textbf{76.6} \\ \hline\hline

\multirow{5}{*}{Cartoon} 
& DeepAll &\textbf{87.4} &67.6 &/ &68.3 &74.5 \\
                               & JiGen    &85.1 &65.5 &/ &65.7 &72.1 \\
                               & GUD      &86.5 &67.2 &/ &68.5 &73.1 \\
                               & M-ADA    & 83.1     & 66.4     &/ &66.3     &71.9      \\ \cline{2-7} 
                               &  NCDG &85.8 &\textbf{71.6} &/ & \textbf{71.9} &\textbf{76.4} \\ \hline\hline

\multirow{5}{*}{Sketch}   
& DeepAll &42.0 &32.2 &54.2 &/  &42.8 \\
                               & JiGen    &47.2 &35.5 &51.8 &/ &44.8 \\
                               & GUD      &32.9 &23.1 &37.5 &/ &31.2 \\
                               & M-ADA    &36.9 &22.0 &42.6 &/ &33.9 \\ \cline{2-7} 
                               &  NCDG &\textbf{47.9} &\textbf{45.6} &\textbf{65.8} &/ &\textbf{53.1} \\ \hline
\end{tabular}
\end{adjustbox}

\label{tab:ssgd-pacs}
\end{table}

 We train NCDG on PACS datasets using backbone architecture Resnet-18 \cite{he2016deep}. We set the learning rate as $0.001$, the batch size as $32$ by training with SGD optimizer. Due to the skip connection existing in Resnet-18, we use the output from the first convolutional layer and the outputs of all the four Resnet blocks before the final fully connected layer for neuron coverage measurement. We consider the original PACS images together with the images processed by the standard augmentation techniques (e.g., image cropping, flipping, jittering), which are widely adopted in ImageNet challenge~\cite{deng2009imagenet} and other domain generalization techniques (e.g., \cite{carlucci2019domain,qiao2020learning}), as the input for fair comparison. The parameters are set as $\lambda= 0.1, t= 0.005, \beta= 0.01$. For baseline implementation, we adopt the open-source codes and report the best performance obtained by tuning the parameters in a wide range. 

The results are reported in Table  \ref{tab:ssgd-pacs}, and we can observe that the proposed method can achieve significantly better performance in all scenarios. Compared with the baseline by directly training on augmented data (i.e., DeepAll), the other single domain generalization methods suffer from performance drop, especially by considering \emph{Sketch} as the source domain. This may originate from the large domain gap between source and target domains. In this case, data augmentation through adversarial training may not be helpful. While JiGen \cite{carlucci2019domain} considers jigsaw puzzle shuffling as a regularization term, it may still suffer from overfitting problem due to the large domain gap, as the fine-grain information learned on the source domain through jigsaw puzzle shuffling still belongs to the source domain, which may not be able to be generalized to the target domain.

\begin{table*}[htbp]
  \centering
\caption{Robustness comparison on \emph{CIFAR-10-C}. We report the classification accuracy (\%) of 19 corruptions under the severest corruption level "5" }    \begin{tabular}{lcccccccccccccc}
\hline
       & \multicolumn{3}{c}{\textbf{Weather}} &  & \multicolumn{5}{c}{\textbf{Blur}}          &  & \multicolumn{4}{c}{\textbf{Noise}}  \\ \cline{2-4} \cline{6-10} \cline{12-15} 
       & Fog            & Snow          & Frost         &  & Zoom          & Defocus       & Glass         & Gaussian       & Motion        & & Speckle       & Shot          & Impulse     & Gaussian \\ \hline
ERM    & 65.9           & 74.3          & 61.6          &  & 60.0          & 53.7          & 49.4          & 30.7           & 63.8          & & 41.3          & 35.4          & 25.7        & 29.0 \\
M-ADA  & 69.4           & 80.6          & 76.7          &  & 68.0          & 61.2          & 61.6          & 47.3           & 64.2          & & 60.9          & 60.6          & 45.2        & 56.9 \\
Augmix & 80.3           & 82.2          & 78.3          &  & 88.0          & 88.9          & 63.5          & 85.2           & 86.4          & & 70.2          & 66.0          & 59.3        & 58.0 \\ \hline
NCDG$_{w/o\ Grad}$   & 80.1           & 81.6          & 80.2          &  & 88.0          & \textbf{89.2} & 66.1          & \textbf{85.3}  & 86.2          & & 72.6          & 69.1          & 62.1        & 63.4 \\
NCDG$_{w/o\ Cov}$   & \textbf{81.5}  & 83.2          & 81.2          &  & 87.7          & 88.8          & 66.3          & 84.5           & 86.2          & & 70.7          & 67.7          & 60.9        & 60.5 \\
NCDG   & 81.1           & \textbf{83.5} & \textbf{82.1} &  & \textbf{88.1} & 89.0          & \textbf{68.0} & 85.2           & \textbf{86.5} & & \textbf{74.7} & \textbf{71.7} & \textbf{66.8}   & \textbf{66.2}\\ \hline
\end{tabular}%
  \label{tab:cifar-c1}%
\end{table*}%
\begin{table*}[htbp]
  \centering
\begin{tabular}{ccccclccccccccccccc}
\cline{6-14}
 &  &  &  &  & \multicolumn{9}{c}{\textbf{Digital}}       &  &  &  &  &  \\ \cline{7-13} 
 &  &  &  &  &        & Jpeg          & Pixelate      & Spatter       & Elastic       & Brightness    & Saturate      & Contrast       & Avg. &  &  &  &  &  \\ \cline{6-14}\
 &  &  &  &  & ERM    & 69.9          & 41.1          & 75.4          & 72.4          & 91.3          & 89.1          & 36.9           & 56.2 &  &  &  &  &  \\
 &  &  &  &  & M-ADA  & 77.1          & 52.3          & 80.6          & 75.6          & 90.8          & 87.6          & 29.7           & 65.6 &  &  &  &  &  \\
 &  &  &  &  & Augmix & 78.5          & 63.0          & 87.4          & 77.3          & 91.6          & 89.8          & 62.0           & 76.6 &  &  &  &  &  \\ \cline{6-14}
 &  &  &  &  & NCDG$_{w/o\ Grad}$   & 78.9 & 63.0          & 88.5          & 77.8          & 92.1          & 90.9          & 62.1           & 77.8 &  &  &  &  &  \\
 &  &  &  &  & NCDG$_{w/o\ Cov}$   & \textbf{79.1}          & 61.3          & \textbf{89.0} & 78.4          & \textbf{93.0}          & \textbf{92.0} & 63.0           & 77.6 &  &  &  &  &  \\
 &  &  &  &  & NCDG   & 78.7          & \textbf{63.4} & 88.6          & \textbf{80.2} & 92.2 & 89.9          & \textbf{69.1}  & \textbf{79.2} &  &  &  &  &  \\
\cline{6-14}
\end{tabular}%
  \label{tab:cifar-c2}%
\end{table*}%

\subsubsection{SSDG (Robustness) Evaluation on CIFAR-10-C}
CIFAR-10-C \cite{hendrycks2019benchmarking} is a typical robustness benchmark consisting of 19 corruptions types with five levels of serveries. The robustness evaluation strategy is similar to SSDG evaluation: each corruption type applied to the original data can be considered as a different domain. We follow the settings of \cite{qiao2020learning} to train our model on CIFAR-10 and evaluate the model on CIFAR-10-C under the highest corruption severity "5". The Wide Residual Network (WRN) \cite{zagoruyko2016wide} with 16 layers and the width 4 is used as the NCDG backbone.  The outputs from the WRN blocks before the final fully connected layer are considered for coverage loss computing, and the hyper-parameters are set as $\lambda= 0.01, t= 0.01, \beta= 0.01$.  Other detailed training settings including the SGD optimizer, the decayed learning rate etc., are the same as \cite{qiao2020learning}. We adopt Augmix for augmentation purpose due to the reason that Augmix is one of the standard techniques for robust deep neural network training \cite{hendrycks2019augmix}.

The results are shown in the  Table. \ref{tab:cifar-c1}.  Based on the results, we have the following observations. First, Augmix can significantly outperform M-ADA, which suggests that the task prior knowledge is important for generalization capability improvement. Second,  both NCDG$_{w/o\ Grad}$ (our proposed method without gradient regularization loss) and NCDG$_{w/o\ Cov}$ (our proposed method without coverage maximization loss) can lead to performance improvement, which verifies the effectiveness of our proposed coverage maximization term and gradient similarity regularization term. Third, by jointly conducting NCDG$_{w/o\ Grad}$ and NCDG$_{w/o\ Cov}$, we can achieve the best performance in most of the cases, further demonstrating the effectiveness of our proposed framework for domain generalization tasks.

\begin{table*}[htbp]
  \centering
  \caption{ 
  Performance comparisons 
  on cross-domain semantic image segmentation.}
  \begin{adjustbox}{max width=\textwidth}
    \begin{tabular}{c|c|c|c|c|c|c}
    \hline
    \multirow{2}[5]{*}{Source} & \multirow{2}[5]{*}{DeepAll} & \multicolumn{2}{c|}{With Target Domain Data } & \multicolumn{3}{c}{W/O Target Domain Data} \\
\cline{3-7}        &       & CycleGAN \cite{dundar2018domain} &  NCDG+CycleGAN    & DR    & DR+ML & NCDG \\
    \hline
    \hline
    GTA5  & 22.7  & 39.6   & 41.4  & 24.9  & 25.3  & \textbf{27.0} \\
    \hline
    SYNTHIA & 18.3  & 27.1  & 30.2  & 19.5  & 19.6  & \textbf{20.1} \\
    \hline
    \end{tabular}%
    \end{adjustbox}
  \label{tab:seg}%
\end{table*}%

\subsubsection{SSDG Evaluation on Cross-Domain Segmentation}
We consider a challenging setting on cross-domain semantic image segmentation task between the synthetic dataset GTA5~\cite{Richter_2016_ECCV}, SYNTHIARAND-CITYSCAPES (SYNTHIA)~\cite{ros2016synthia} and real-world dataset CITYSCAPES~\cite{cordts2016cityscapes}, where the synthetic dataset is used as source domain and the CITYSCAPES dataset is used as unseen target domain.  In particularly,  we adopt DRN-C-26 \cite{yu2017dilated} as the backbone, crop the image with the size 600 $\times$ 600, and apply SGD with learning rate $0.001$ and momentum 0.9 with a batch size of 8 for training purpose. We follow \cite{dundar2018domain} by applying domain randomization (DR) for augmentation purpose. We report mean intersection-over-union (mIoU)  for evaluation and the results are shown in Table \ref{tab:seg}. As we can see, our proposed method can achieve consistently better performance on segmentation task where unlabeled target domain data. It is worth noting that we also conduct adversarial learning \cite{qiao2020learning,volpi2018generalizing} for data augmentation but find it cannot achieve desired performance compared with DR. We conjecture the reason that adversarial learning may not handle the large domain shift between synthetic data and real-world data.  Furthermore, we consider meta-learning (ML), which was adopted in \cite{qiao2020learning}, as another baseline based on the original and augmented data through DR. Based on the results, we find our proposed NCDG can also achieve better performance,  which further justifies the effectiveness of our proposed method. We then consider the domain adaptation setting where the prior knowledge in target domain is available. Particularly, we follow \cite{dundar2018domain} by applying CycleGAN for style transfer, and subsequently apply our proposed method based on the new data. As we can see, our method is also effective by combining with domain adaptation techniques, which further justifies the effectiveness of our proposed method.

Last but not the least, we also consider more strong competitors to justify the effectiveness of our proposed method. Specifically, we apply our proposed method to the Deeplab v3+ \cite{chen2018encoder}, Deeplab v3+ with IBN layer \cite{pan2018two} and RobustNet \cite{choi2021robustnet} (i.e., Deeplab v3+ with ISW) with Resnet50 as the backbone (as suggested in \cite{choi2021robustnet}) by considering GTAV as source domain and CityScape as target domain. The results are shown below. As we can observe, our proposed method can consistently achieve the best performance in all scenarios, which show the effectiveness of our proposed method.  It is also worth noting that our proposed method can achieve the state-of-the-art performance by comparing with the recent method Robostnet. 

\makeatletter\def\@captype{table}\makeatother

\begin{table}[h]
\begin{center}
\caption{Performance comparisons 
  on cross-domain semantic image segmentation on DeepLabv3+ with Resnet50 as backbone. }
\begin{adjustbox}{max width=\columnwidth}
\begin{tabular}{cccc}
\hline
Method  &Baseline  & IBN  & ISW   \\ \hline\hline
w/o coverage &29.7   &33.9 &36.6      \\
w/ {coverage}      &\textbf{34.7}   &\textbf{35.6} &\textbf{37.2}     \\
\hline
\end{tabular}
\end{adjustbox}
\end{center}
\label{tab:table-l1}
\end{table}
~\\

\subsubsection{Component Analysis}
To investigate the specific role of each NCDG objective loss component, besides the results we show in Table \ref{tab:cifar-c1}, we also conduct detailed ablation experiments on PACS datasets using SSDG settings based on Resnet-18. Specifically, we evaluate the performance of NCDG without the gradient regularization term as well as NCDG without coverage maximization. The latter is achieved by setting neuron coverage weight $\lambda$ to zero to eliminate the influences of neuron coverage loss. The results are shown in Table \ref{tab:pacs-abla}. 

As we can see, generally, both coverage loss and similarity regularization contribute to the final performance by comparing with the DeepAll baseline in Table \ref{tab:ssgd-pacs}. We also observe that, in most of the cases, the performance can be further boosted by combining the two together, which further justify the effectiveness of our proposed method. However, in some other cases, our proposed gradient regularization method may not be helpful, especially when using ``Photo" as the target domain. We conjecture that there are two possible reasons.
\begin{itemize}
    \item Compared with the target domain, the sample diversity on source domain as well as its corresponding augmented domain may not be that high. Therefore, applying gradient regularization on source domain may not be able to capture reliable decision behavior of the target.
    
    \item When using Art-Painting and Cartoon as source domains, we can already achieve a relatively high performance by using Photo as target domain with coverage maximization only. Therefore, applying gradient regularization may lead to over-fitting problem on source domain. 
\end{itemize}

It is also interesting to analyze impact of applying neuron coverage loss at different layers of the network. To this end, we use sketch as source domain and the others as target domains. In our original experiment setting, we use the outputs from the first convolutional layer and all the four Resnet blocks for coverage loss computing. To further understand the relationship between the neuron location and final performance, we conduct ablation studies by considering different number of layer outputs, where $conv_{0}$ denotes the output of the first convolutional layer, and $conv_{0,1,..,k}$ denotes the output of the first convolutional layer along with the outputs of Resnet block(s) from $1$ to $k$. The results are shown in Table \ref{tab:table-layer}. As we can see, the performances can be variant by considering different layers. In object recognition task based on PACS, we observe that, in most of the cases, better performance can be achieved by considering more neurons for coverage loss computing.

Besides the loss components, we are also interested in whether our proposed method can be applied to the network with different activation functions. To this end, we further consider PReLU and ELU on the single domain generalization task based on PACS dataset. In particular, as shown in Table \ref{tab:table-activation}, Sketch domain is used as the source domain and the other domains are treated as target. As we can see, our proposed method can consistently achieve better performance when using PReLU and ELU as the activation functions.

\begin{table}[]
\caption{Component analysis on PACS dataset. }
\begin{adjustbox}{max width=\columnwidth}
\begin{tabular}{c|lcccc|c}
\hline
Source Domain                  & Method   & Photo & Art\_painting & Cartoon & Sketch & Avg. \\  \hline\hline
\multicolumn{7}{c}{\textbf{Resnet-18}}  \\ \hline
\multirow{3}{*}{Photo}         
& NCDG$_{w/o\ Grad}$    &/ &65.1 &25.9 &39.9  &43.6 \\
& NCDG$_{w/o\ Cov}$    &/ &67.0 &29.0 &45.9  &47.3 \\
& NCDG &/ &\textbf{68.8} &\textbf{29.8}- &\textbf{48.6} &\textbf{49.0} \\ \hline\hline

\multirow{3}{*}{Art\_painting} 
                               & NCDG$_{w/o\ Grad}$    &\textbf{96.4} &/ &65.7 &59.0 &73.7 \\
                               & NCDG$_{w/o\ Cov}$    &93.8 &/ &67.4 &63.7 &75.0 \\
                               & NCDG &95.0 &/ &\textbf{68.6} &\textbf{66.4} &\textbf{76.6} \\ \hline\hline

\multirow{3}{*}{Cartoon}       
                               & NCDG$_{w/o\ Grad}$    &\textbf{88.6} &70.9 &/ &69.3 &76.3 \\
                               & NCDG$_{w/o\ Cov}$    &86.2 &69.3 &/ &\textbf{72.6} &76.0 \\
                               & NCDG &85.8 &\textbf{71.6} &/ &71.9 &\textbf{76.4} \\ \hline\hline

\multirow{3}{*}{Sketch}        
                               & NCDG$_{w/o\ Grad}$   &\textbf{49.2} &35.8 &51.9 &/ &45.6 \\
                               & NCDG$_{w/o\ Cov}$    &45.9 &36.8 &53.4 &/ &45.4 \\
                               & NCDG &47.9 &\textbf{45.6} &\textbf{65.8} &/ &\textbf{53.1} \\ \hline
\end{tabular}
\end{adjustbox}

\label{tab:pacs-abla}
\end{table}

\makeatletter\def\@captype{table}\makeatother
\begin{table}[h]
\caption{Layer analysis of Resnet-18 on \emph{PACS} (Sketch as the source domain).}
\begin{center}
\begin{adjustbox}{max width=\columnwidth}
\begin{tabular}{lccccc} \hline
 Layers    & Photo  & Art & Cartoon  & Avg.  \\ \hline
$conv_{0}$          & 44.9 & 33.1   & 62.2  & 46.7 \\
$conv_{0,1}$        & 46.8 & 35.5   & 63.3  & 48.5 \\ 
$conv_{0,1,2}$      & 45.3 & 44.2   & 64.5  & 51.4 \\
$conv_{0,1,2,3}$    & \textbf{48.7} & 44.7   & 63.9  & 52.4 \\ \hline
$conv_{0,1,2,3,4}$  & 47.9 & \textbf{45.6}   & \textbf{65.8}  & \textbf{53.1} \\ \hline

\hline

\end{tabular}
\end{adjustbox}
\end{center}

\label{tab:table-layer}
\end{table}

\makeatletter\def\@captype{table}\makeatother
\begin{table}[h]
\caption{SSDG comparison between different activation functions of Resnet-18 on \emph{PACS} (Sketch as the source domain).}
\begin{center}
\begin{adjustbox}{max width=\columnwidth}
\begin{tabular}{lccccc} \hline
Activation Func   & Photo  & Art & Cartoon  & Avg.  \\ \hline
DeepAll$_{PReLU}$  & 44.7 & 34.3 & 52.1 & 43.7 \\
NCDG$_{PReLU}$     & 45.0 & 36.5 & 61.7 & 47.7 \\ \hline
DeepAll$_{ELU}$    & 43.7 & 33.5   & 52.1  & 43.1 \\
NCDG$_{ELU}$       & 45.0 & 37.8   & 60.9 & 47.9 \\ \hline
DeepAll$_{ReLU}$   & 42.0 & 32.2   & 54.2  & 42.8 \\
NCDG$_{ReLU}$      & \textbf{47.9} & \textbf{45.6}   & \textbf{65.8} & \textbf{53.1} \\ \hline

\hline

\end{tabular}
\end{adjustbox}
\end{center}

\label{tab:table-activation}
\end{table}

It is worth noting that feature selection can also benefit generalization capability \cite{gui2016feature}, and our proposed method is conceptually different with the feature selection as we require all neurons to be activated by at least one training sample. Therefore, we are also interested in the superiority of our proposed coverage loss against feature selection based techniques. To this end, we replace our coverage loss with L1 regularization \cite{gui2016feature} term on hidden layers which are also considered for the coverage loss. In particular, we consider the single domain generalization task by using PACS benchmark, where Sketch is used as the source domain, and the others are treated as target domains. We tune the hyperparameter of L1 regularization in a wide range to report the best performance we can obtain. As we can see from Table 7, L1 regularization can generally achieve better performance compared with DeepAll baseline, which shows that feature selection can help with generalization performance. Nevertheless, our proposed method can achieve much better performance compared with results using L1 regularization, which shows the effectiveness of our proposed method. 

\begin{table}[h]
\begin{center}
\caption{SSDG Comparison between NCDG proposed loss and L1 regularization loss on PACS dataset (Sketch domain as the source) }
\begin{adjustbox}{max width=\columnwidth}
\begin{tabular}{lcccc}
\hline
Method  &Photo  &Art  &Cartoon  &Avg.  \\ \hline\hline
DeepAll &42.0   &32.2 &54.2     &42.8 \\
L1      &41.5   &32.7 &64.0     &46.1 \\ \hline
NCDG    &\textbf{47.9} &\textbf{45.6} &\textbf{65.8} &\textbf{53.1} \\
\hline
\end{tabular}
\end{adjustbox}
\end{center}
\label{tab:table-lone}
\end{table}

Last but not the least, it is beneficial to look at divergence between source and target domain. To this end, we calculate the Kullback–Leibler (KL) divergence between source and target domain by using ``sketch" as source domain and the other three as target domain. We use the output of the last Resnet block as the feature space. The results are shown as follow. We can draw the conclusion that the distribution distance between source and target domain based on KL divergence can be decreased when applying our method compared with ``DeepAll".

\makeatletter\def\@captype{table}\makeatother

\begin{table}[h]            
\begin{center}
\caption{Kullback–Leibler divergences between the source domain: Sketch, with the other three target domains of DeepAll and NCDG models.}
\begin{adjustbox}{max width=\columnwidth}
\begin{tabular}{lccc}
\hline
Method   & Photo  & Art & Cartoon   \\ \hline\hline
DeepAll  &  0.0044 & 0.0043 & 0.0027  \\
NCDG & \textbf{0.0030} &\textbf{0.0022}   & \textbf{0.0018} \\
\hline

\end{tabular}
\end{adjustbox}

\label{tab:table-kl}
\end{center}
\end{table}

\begin{figure}[]
	\centering
	\subfigure[]{\includegraphics[width=0.3\columnwidth]{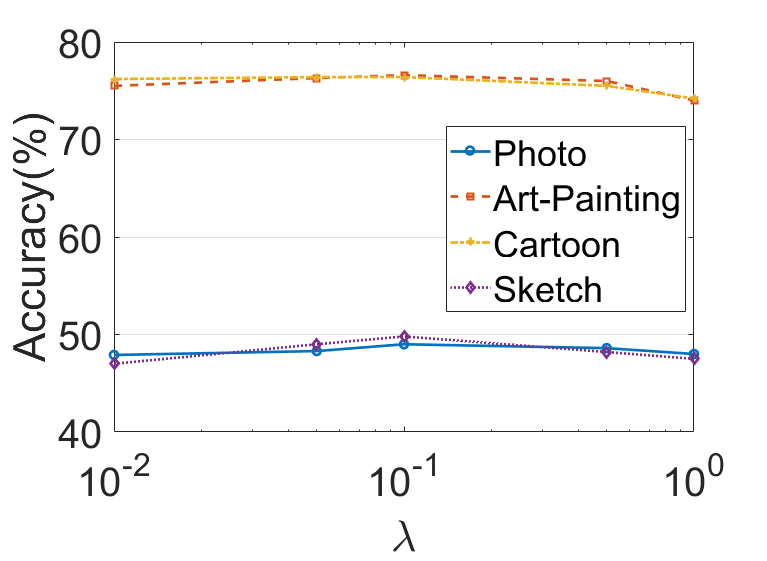}}
	\subfigure[]{\includegraphics[width=0.3\columnwidth]{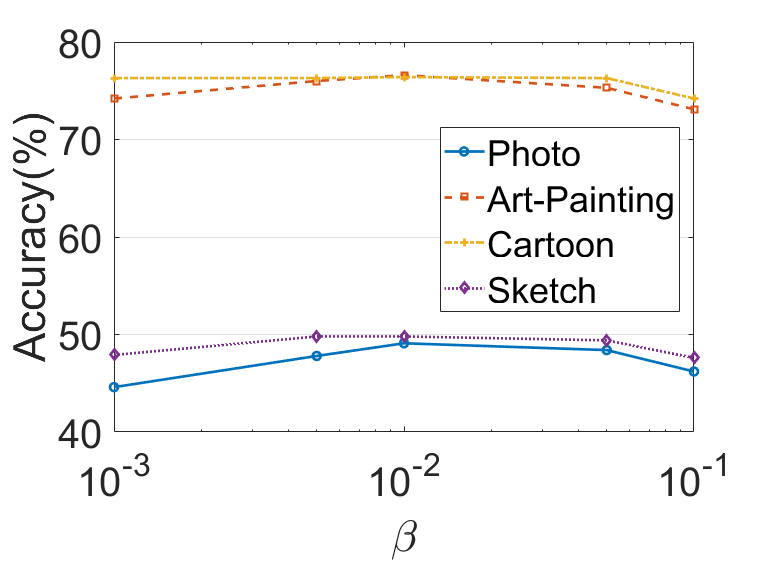}}
	\subfigure[]{\includegraphics[width=0.3\columnwidth]{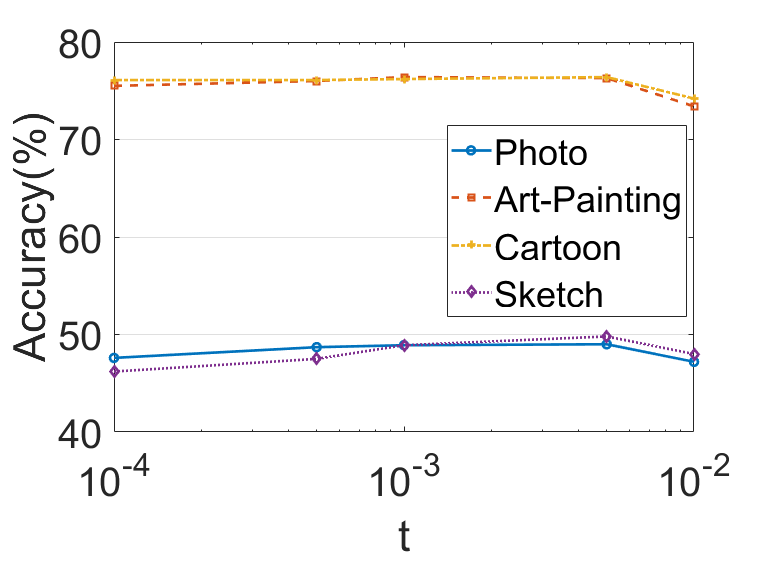}}
	\caption{Parameter sensitivity analysis by varying (a) $\lambda$, (b) $\beta$, (c) ${t}$. Each curve denotes the performance by considering the domain shown in legend as source domain. The performance is reported by averaging the results on other three domains.  }\label{fig:parameter_dg}
	\vspace{-10pt}
\end{figure}

 \subsubsection{Parameter Sensitivity Analysis }
qWe conduct parameter sensitivity study on PACS with Resnet-18 based on SSGD setting. More specifically, we use one domain from PACS as source and report the average classification accuracy on the other three domains. The results are shown in Fig.~\ref{fig:parameter_dg} by varying the parameters. As we can see, the final performance is stable based on different parameters, which shows the robustness of our proposed method. However, extreme parameters may lead to significant performance drop. For example, when lowering $\beta$ which controls ``similarity", the performance decreases as the effectiveness of similarity regularization can be reduced by lowering $\beta$. On the other hand, the performance also drops when $\beta>=0.05$, 
as in this scenario the objective over-fits to the similarity regularization term.



\subsection{Multi-Source Domain Generalization}
We further validate the proposed method NCDG with a more general setting, where multiple source domains are available during training, i.e., Multi-Source Domain Generalization ({MSDG}). We treat the training data as a compound source domain where domain knowledge is not available in the model training of NCDG. The evaluations are conducted on PACS as well as two other benchmark datasets Office-Home \cite{venkateswara2017deep} and VLCS \cite{torralba2011unbiased}. 





It is worth noting that the aforementioned benchmark datasets contain different numbers of categories ranging from 5 to 65. Such difference renders a well-established test-bed to verify the scalability of NCDG as diversities of object categories are considered.  Following the existing setting, we apply the leave-one-out strategy for training and evaluation, such that one domain is selected as 
the target while the remaining domains are treated as source. We also follow the same data augmentation and neuron coverage computing strategy  as PACS based on SSDG setting. 


We compare NCDG against several state-of-the-art domain generalization baselines, including TF \cite{li2017deeper}, MMD-AAE \cite{li2018domain}, D-SAM \cite{d2018domain}, JiGen \cite{carlucci2019domain}, L2A-OT \cite{zhou2020deep}, DDAIG \cite{zhou2020learning}, EISNet \cite{wang2020learning}, DSON \cite{seo2020learning}, InfoDrop \cite{shi2020informative} and RSC \cite{huang2020self}. 
In particular, both L2A-OT and DDAIG require specific  prior domain knowledge regarding the source training data for adversarial domain generation.



\subsubsection{MSDG Evaluation on Office-Home}
Following the setting of existing DG methods \cite{carlucci2019domain, zhou2020deep}, for each domain, we randomly select 90\% for training and 10\% for validation, using ImageNet \cite{deng2009imagenet} pre-trained Resnet-18 \cite{he2016deep} as the backbone model with training setting as PACS SSDG experiment. The introduced hyper-parameters in NCDG are set as $\lambda= 1, t= 0.005, \beta= 0.01$ for training purpose. 

\begin{table}[]
\caption{MSDG classification accuracy (\%) on Office-Home datasets. Each column indicates the results on a given target domain. }
\begin{adjustbox}{max width=\columnwidth}
\begin{tabular}{lcccc|c}
\hline
\textbf{Office-Home} &\textbf{Art} &\textbf{Clipart} &\textbf{Product} &\textbf{Real-World} &\textbf{Avg.}\\ \hline \hline
DeepAll &55.6 &42.4 &70.3 &70.9 &59.8 \\
D-SAM   &58.0 &44.4 &69.2 &71.5 &60.8          \\
JiGen   &53.0 &47.5 &71.5 &72.8 &61.2          \\
L2A-OT  &\textbf{60.6} &50.1 &74.8 &\textbf{77.0} &65.6          \\
DDAIG   &59.2 &52.3 &74.6 &76.0 &65.5          \\
RSC     &58.4 &47.9 &71.6 &74.5 &63.1          \\

DSON     &59.4          &45.7 &71.8    &74.7  &62.9 \\
\hline
NCDG &59.8 &\textbf{53.1} &\textbf{75.3} &76.3 &\textbf{66.1} \\ \hline
\end{tabular}
\end{adjustbox}

\label{tab:officehome-msdg}
\end{table}


The results are shown in Table \ref{tab:officehome-msdg}. As we can see, by comparing with the baseline methods D-SAM and JiGen which do not require domain knowledge during training stage, we can achieve significantly improvement in all scenarios.  On the other hand, by comparing with the baselines where domain knowledge is utilized, we can also achieve competitive performance and achieve the best performance on average, indicating the effectiveness of our proposed method.

\subsubsection{MSDG Evaluation on VLCS}
 Following \cite{li2018domain, carlucci2019domain}, we choose ImageNet pre-trained AlexNet \cite{krizhevsky2012imagenet} as the backbone for evaluation, with all images resized to $225 \times 225$. For parameter selection, we choose $\lambda=0.1,t=0.001,\beta=0.01$. We also use SGD for model training and the setting is the same as Office-Home.

The results are reported in Table \ref{tab:vlcs-msdg}. As we can see, we can achieve competitive performance when using \emph{Caltech}, \emph{Pascal} and \emph{Sun} as target domains, and significant improvement can be achieved by evaluating on \emph{LabelMe}. Overall, we can achieve the best performance on average by comparing with all other baselines. 



\begin{table}[]
\caption{ MSDG classification accuracy (\%) on VLCS. Each column indicates the results on a given target domain.
}
\begin{adjustbox}{max width=\columnwidth}
\begin{tabular}{lcccc|c}
\hline
\textbf{VLCS}   &\textbf{Caltech} &\textbf{LabelMe} &\textbf{Pascal} &\textbf{Sun} &\textbf{Avg.}\\ \hline \hline
DeepAll &85.7 &61.3 &62.7 &59.3 &67.3 \\
TF &93.6 &63.4 &70.0 &61.3 &72.1 \\ 
MMD-AAE &94.4 &62.6 &67.7 &64.4 &72.3 \\ 
D-SAM &91.8 &57.0 &58.6 &60.8 &67.0 \\ 
JiGen &96.9 &60.9 &70.6 &64.3 &73.2 \\
RSC   & \textbf{97.6} &61.9 &\textbf{73.9} &68.3 &75.4 \\
EISNet   &97.3          &63.5          &69.8    &68.0  &74.7 \\
\hline
NCDG & 97.2        &\textbf{67.6} & 70.7 & \textbf{68.7} & \textbf{76.1} \\ \hline
\end{tabular}
\end{adjustbox}

\label{tab:vlcs-msdg}
\end{table}


\begin{table}[]
\caption{MSDG classification accuracy (\%) on PACS. Each column indicates the results on a given target domain.
}
\begin{adjustbox}{max width=\columnwidth}
\begin{tabular}{lcccc|c}
\hline
\textbf{PACS}  &\textbf{Photo} &\textbf{Art painting} &\textbf{Cartoon} &\textbf{Sketch} &\textbf{Avg.} \\ \hline \hline

\multicolumn{6}{c}{\textbf{AlexNet}}  \\ \hline
DeepAll &87.7 &63.3 &63.1 &54.1 &67.1 \\
D-SAM   &85.6 &63.9 &69.4 &64.7 &71.2 \\
JiGen   &89.0 &67.6 &71.7 &65.2 &73.4 \\ 
RSC     &90.9 &\textbf{71.6} &\textbf{75.1} &66.6 &76.1 \\
EISNet   &\textbf{91.2}          &70.4          &71.2    &70.3  &75.9 \\
\hline
NCDG    & 89.0        & 68.9 & 74.7 & \textbf{72.9} & \textbf{76.4} \\ \hline

\multicolumn{6}{c}{\textbf{Resnet-18}}  \\ \hline
DeepAll &94.3          &77.4          &75.7    &69.6  &79.2 \\
D-SAM   &95.3          &77.3          &72.4    &77.8  &80.7 \\
JiGen   &96.0          &79.4          &75.3    &71.4  &80.5 \\ 
L2A-OT  &\textbf{96.2} &83.3          &78.2    &73.6  &82.8 \\
DDAIG   &95.3          &84.2          &78.1    &74.7  &83.1 \\
RSC     &96.0          &83.4          &80.3    &80.9  &85.2 \\
EISNet   &95.9          &81.9          &76.4    &74.3  &82.2 \\
InfoDrop &96.1          &80.3          &76.5    &76.4  &82.3 \\
DSON     &95.9          &\textbf{84.7} &77.7    &\textbf{82.3}  &85.1 \\
\hline
NCDG    &95.4          &82.3          &\textbf{82.3} &82.1 &\textbf{86.2} \\ \hline
\end{tabular}
\end{adjustbox}

\label{tab:pacs-msdg}
\vspace{-10pt}
\end{table}


\subsubsection{MSDG Evaluation on PACS}
To demonstrate the model-agnostic generalization ability that NCDG brings, we follow \cite{carlucci2019domain} to train NCDG on PACS using backbone architectures Resnet-18 and AlexNet individually. For Resnet-18, we set the parameters in NCDG as $\lambda=1, t=0.005, \beta=0.01$, and $\lambda=0.1, t=0.001,  \beta=0.01$ are used for AlexNet. Other training details are the same as those for Office-Home based on ResNet-18 and VLCS based on AlexNet.

The results in Table \ref{tab:pacs-msdg} reveal that our proposed method can achieve better performance when considering \emph{Cartoon} and \emph{Sketch} as the target domains, and achieve competitive performance when using \emph{Photo} and \emph{Art Painting} as target domains by comparing the baselines leveraging domain information.  We can also achieve the best performance on average, which further shows the effectiveness of our proposed method.



\begin{figure}[t]
\centering
\includegraphics[width=0.9\columnwidth]{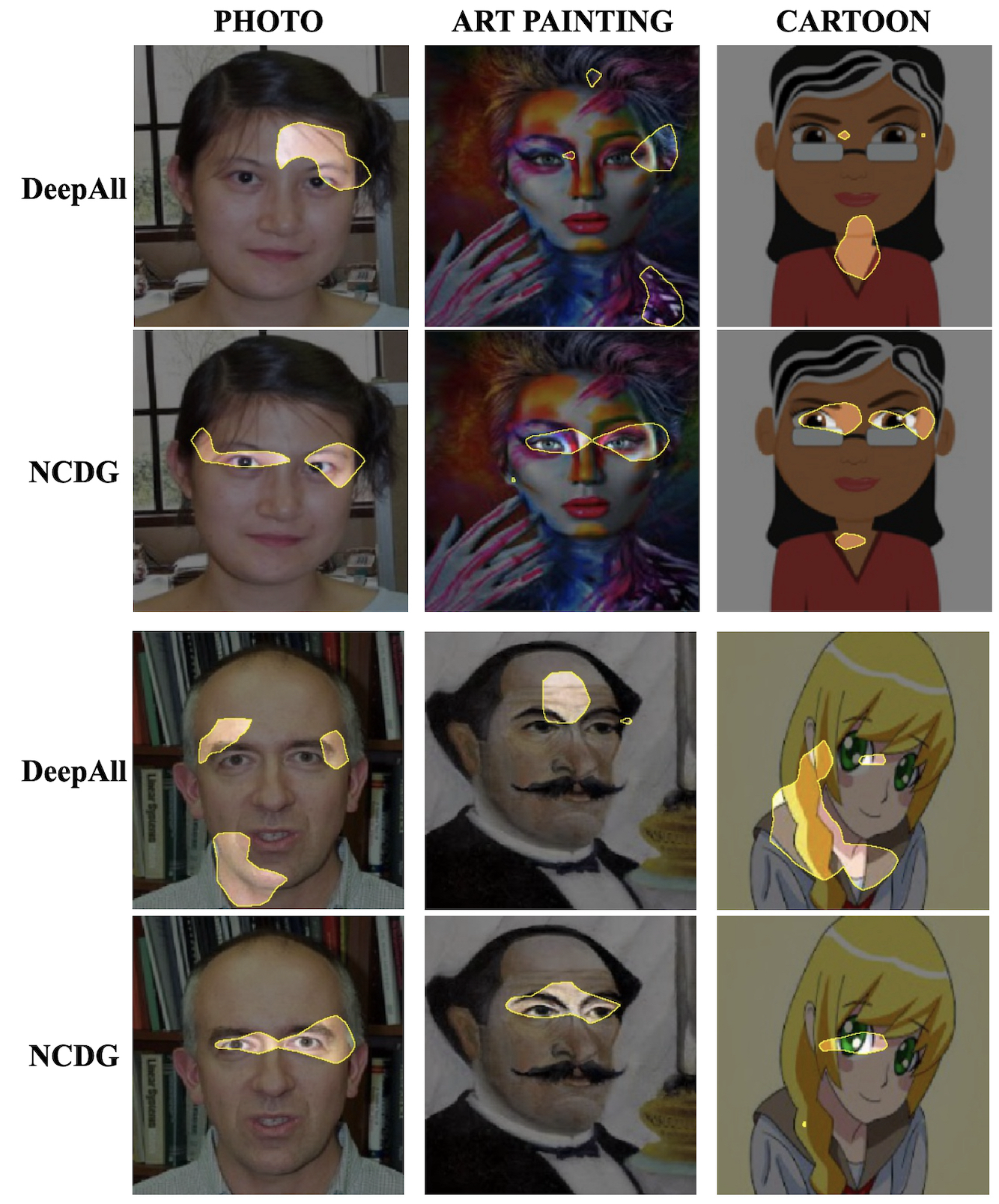}
\caption{Visualization results of network dissection. Each column indicates images from one target domain and each row shows the visualization result of unit 170 by either DeepAll model or our proposed NCDG.}
\label{fig:vis}
\end{figure}

\subsection{Connection with Network Dissection}
Here we provide in-depth analyses regarding the effectiveness of our proposed method. In particularly, we aim to justify our motivation based on network dissection \cite{bau2020understanding}, which is to identify the semantics information of individual hidden units (i.e., channel) of the image classification network. 

Recall that our proposed method is built upon the assumption that an out-of-distribution sample for evaluation may trigger the inactive neuron, which further leads to misclassification behavior. To better understand how this behavior is triggered, we can examine the unit of network trained by both DeepAll strategy (i.e., directly training with cross-entropy loss) and our proposed algorithm. Specifically, we consider using PACS dataset based on single domain generalization setting trained with Resnet-18 as an example, where the network is trained based on Sketch domain and evaluated on the other three domains.  We choose two images belonging to the ``person" category, which are misclassified by DeepAll but can be correctly predicted by our proposed method, from each target domain. We further visualize unit 170 from the output of block 3 of Resnet-18 by feeding the images to the network, as we empirically find that unit 170 is not activated if we simply train Resnet-18 based on Sketch domain by using DeepAll strategy. By contrast, the unit can be activated by using our proposed method. The visualization results are shown in Fig.~\ref{fig:vis}.

As we can observe, all testing samples can activate the unit 170 of Resnet-18 block 3. No semantic information is observed from unit 170 when directly training the network with DeepAll, thus it is reasonable that the network predicts wrongly as the activated unit 170 triggers unexpected behavior. We can also observe that the concept of ``eye" can be consistently extracted by our proposed method across samples, and such concept has a strong connection with the ``person" category, which further 
drives the network to predict correctly. Such observation justifies our motivation that maximizing the neuron coverage facilitates the reduction on the number of possible defects introduced by out-of-distribution sample.  On the other hand, our observation is also consistent with the finding in \cite{bau2020understanding} that the classification performance of a certain category can be explained by the units that identify visual concepts of this class.

\begin{figure}[!t]
\centering
\includegraphics[width=0.9\columnwidth]{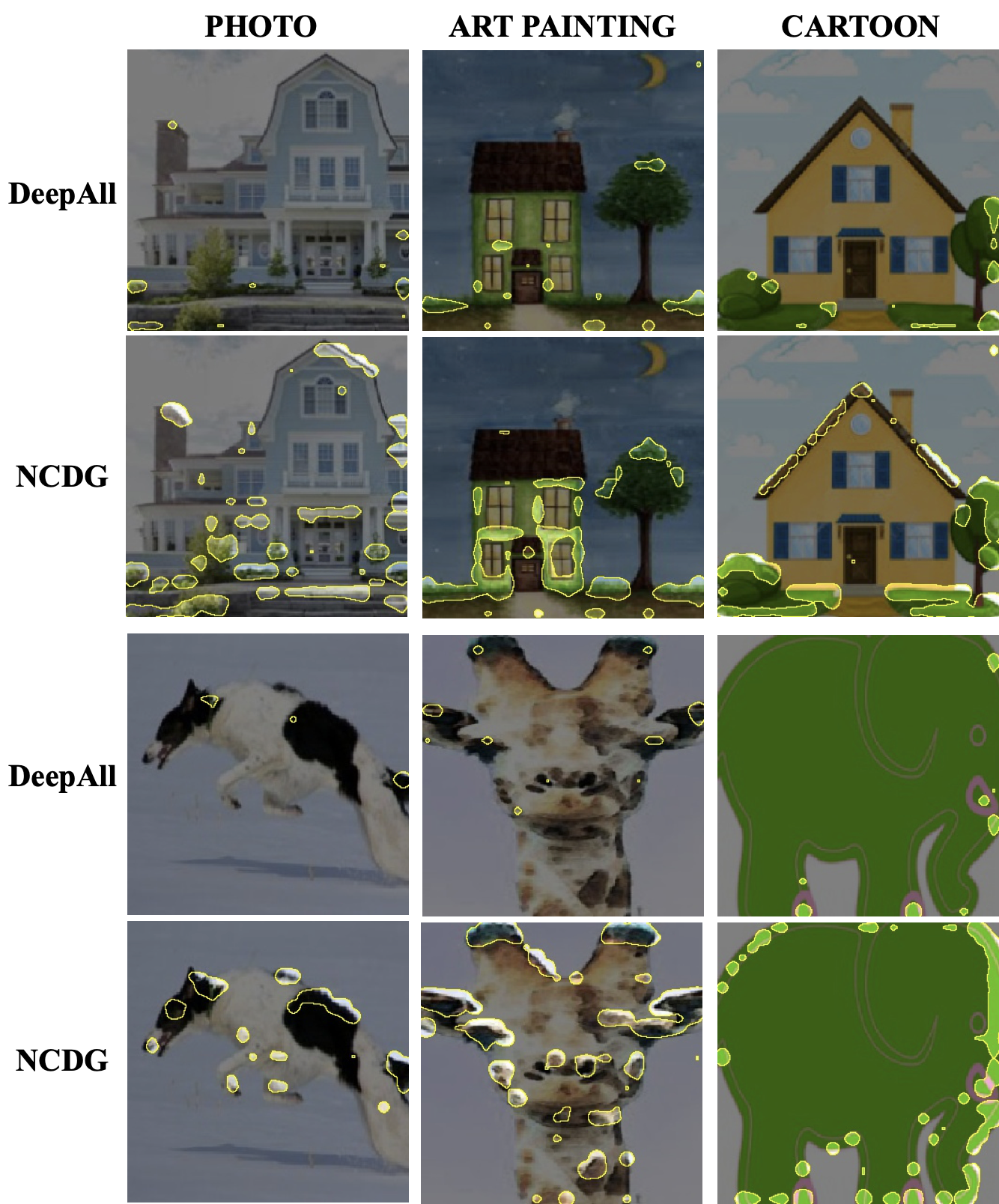}
\caption{Visualization results of network dissection. Each column indicates images from one target domain and each row shows the visualization result of ResNet-18 block 2 unit 20 by either DeepAll model or our proposed NCDG.}
\label{fig:vis2}
\end{figure}

Besides focusing on the output of ResNet-18 block 3 which captures semantic information, we are also interested in the low-level block output. To this end, we visualize the unit 20 of ResNet-18 block 2 by considering sketch as source domain and the remaining ones as target domain based on different categories.
Based on the results in  Fig.~\ref{fig:vis2}, we observe that our proposed method can explore some edge and contour information, which can benefit object recognition. Again, it can be difficult for us to observe useful information extracted by DeepAll baseline. Thus, it is reasonable that the network predicts wrongly if the given neuron is activated by out-of-distribution samples.

\section{Conclusions}
In this paper, we propose to improve the generalization capability of DNN from the perspective of neuron coverage maximization. By modeling the DNN as a program, we propose to maximize the neuron coverage (i.e., control flow) of DNN with the gradient (i.e., data flow) similarity regularization  between the original data and the augmented data during the training stage, such that the trained DNN can be better generalized to the out-of-distribution samples. Extensive experiments on various domain generalization tasks 
verify the effectiveness of our proposed method. 



{
\small
\bibliographystyle{IEEEtran}
\bibliography{bare_jrnl_compsoc}
}


%

\end{document}